\documentclass{article} 

\usepackage{nips14submit_e,times}
\usepackage{moreverb,url}
\usepackage[pdftex]{graphicx}
\usepackage{amsmath}
\usepackage{amssymb}
\usepackage[justification=centering]{caption}
\usepackage{graphicx}

\title{Diversity Regularized Adversarial Learning}

\author{Babajide~O. Ayinde\thanks{B.~O.~Ayinde is with the Department
		of Electrical and Computer Engineering, University of Louisville, Louisville,
		KY, 40292 USA (e-mail: babajide.ayinde@louisville.edu).}
    , Keishin Nishihama\thanks{Keishin Nishihama (e-mail: keishinkickback@gmail.com).}
    , and Jacek~M.~Zurada \thanks{J.~M.~Zurada is with the Department
		of Electrical and Computer Engineering, University of Louisville, Louisville,
		KY, 40292 USA, and also with the Information Technology Institute, University of Social Science,\L \'{o}dz 90-113, Poland (Corresponding author, e-mail: jacek.zurada@louisville.edu). This work was supported by the NSF under grant 1641042.}
}

\DeclareMathOperator{\E}{\mathbb{E}}

\nipsfinalcopy 

\begin{document}

\maketitle

\begin{abstract}
The two key players in Generative Adversarial Networks (GANs), the discriminator and generator, are usually parameterized as deep neural networks (DNNs). On many generative tasks, GANs achieve state-of-the-art performance but are often unstable to train and sometimes miss modes. A typical failure mode is the collapse of the generator to a single parameter configuration where its outputs are identical. When this collapse occurs, the gradient of the discriminator may point in similar directions for many similar points. We hypothesize that some of these shortcomings are in part due to primitive and redundant features extracted by discriminator and this can easily make the training stuck. We present a novel approach for regularizing adversarial models by enforcing diverse feature learning. In order to do this, both generator and discriminator are regularized by penalizing both negatively and positively correlated features according to their differentiation and based on their relative cosine distances. In addition to the gradient information from the adversarial loss made available by the discriminator, diversity regularization also ensures that a more stable gradient is provided to update both the generator and discriminator. Results indicate our regularizer enforces diverse features, stabilizes training, and improves image synthesis.
\end{abstract}


\section{Introduction}
Convolutional neural networks (CNNs) have become the powerhouse for tackling many image processing and computer vision tasks. By design, CNNs learn to automatically optimize a well-defined objective function that quantifies the quality of results and their performance on the task at hand. As shown in previous studies \cite{Isola_2017_CVPR}, designing effective loss functions for many image prediction problems is daunting and often requires manual effort and in-depth experts' knowledge and insights. For instance, naively minimizing the Euclidean distance between predicted and ground truth pixels have shown to result in blurry outputs since the Euclidean distance is minimized by averaging all conceivable outputs \cite{Isola_2017_CVPR,pathak2016context,zhang2016colorful,ayinde2018deep}. One plausible way of training models with high-level objective specifications is by allowing CNNs to automatically learn the appropriate loss functions that satisfy these desired objectives. One of such objectives could be as simple as asking the model to make the output not distinguishable from the groundtruth.\\
\indent
As established in \cite{goodfellow2014generative,radford2015unsupervised,Isola_2017_CVPR,salimans2016improved}, GANs are trained to automatically learn an objective function using a discriminator network to classify if its input is real or synthesized while simultaneously training a generative model to minimize the loss. In GAN framework, both the discriminator and generator aim to minimize their own loss and the solution to the game is the Nash equilibrium where neither player can independently improve their individual loss \cite{goodfellow2014generative,kurach2018gan}. This framework can also be interpreted from the viewpoint of a statistical divergence minimization between the learned model distribution and the true data distribution \cite{nowozin2016f,arjovsky2017wasserstein,mao2017least}.\\
\indent
Even though GANs have resulted in new and interesting applications and achieved promising performance, they are still hard to train and very sensitive to hyperparameter tuning. A peculiar and common training challenge is the performance control of the discriminator. The discriminator is usually inaccurate and unstable in estimating density ratio in high dimensional spaces, thus leading to situations where the generator finds it difficult to model the multi-modal landscape in true data distribution. In the event of total disjoint between the supports of model and true distributions, a discriminator can trivially distinguish between model distribution and that of true data \cite{arjovsky2017towards}, thus leading to situations where generator stops training because the derivative of the resulting discriminator with respect to the input has vanished. This problem has seen many recent works to come up with workable heuristics to address many training problems such as mode collapse and missing modes. \\
\indent
We argue in line with the hypothesis that some of the problems associated with the training of GANs are in part due to lack of control of the discriminator. In light of this, we propose a simple yet powerful diversity regularizer for training GANs that encourages the discriminator to extract near-orthogonal filters. The problem abstraction is that in addition to the gradient information from the adversarial loss made available by the discriminator, we also want the GAN system to benefit from extracting diverse features in the discriminator. Experimental results consistently show that, when correctly applied, the proposed regularization enforces diverse features in the discriminator and better stabilize the GAN training with mostly positive effects on the generated samples.\\
\indent
The contribution of this work is two-fold: (i) we propose a new method to regularize adversarial learning by inhibiting the learning of redundant features and availing a stable gradient for weights updates during training and (ii) we show that the proposed method stabilizes the adversarial training and enhances the performance of many state-of-the-art methods across many benchmark datasets. The rest of the paper is structured as follows: Section II highlights the state-of-the-art and Section III discusses in detail the formulation of diversity-regularized adversarial learning. Section IV discusses the detailed experimental designs and presents the results. Finally, conclusions are drawn in Section V.
\section{Related Work}
\begin{figure*}[htb!]
  \centering
  \captionsetup{justification=centering}
  \includegraphics[scale=0.6]{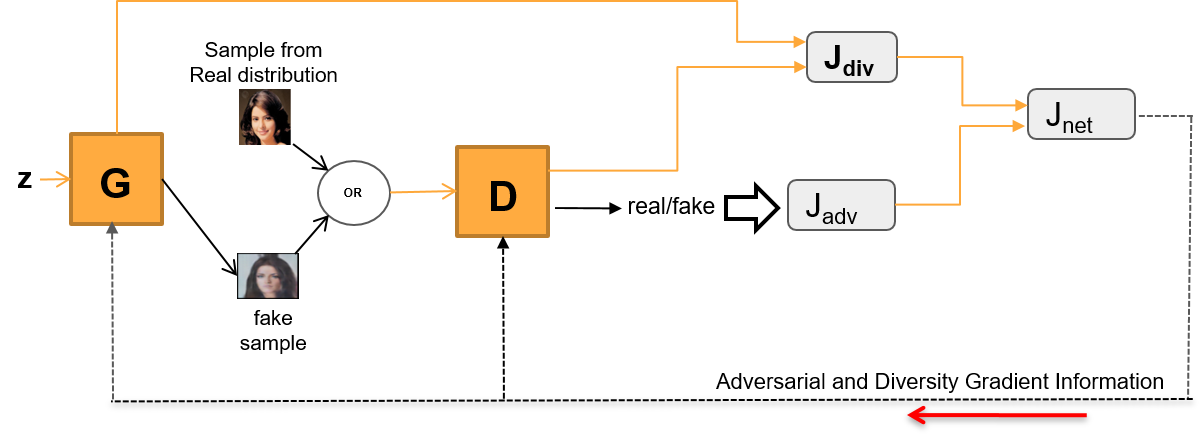}
  \caption{Schema of Diversity Regularized Adversarial Learning (DiReAL)}\label{gan_image}
\end{figure*}
As originally introduced in \cite{goodfellow2014generative}, GANs consist of generator and the discriminator that are parameterized by deep neural networks and are capable of synthesizing interesting local structure on select datasets. The representation capacity of original GAN was extended in conditional GANs \cite{mirza2014conditional} by incorporating an additional vector that enables the generator to synthesize samples conditioned on some useful information. This extension has motivated several conditional variants of GAN in diverse applications such as edge map \cite{zhu2017unpaired,isola2017image}, image synthesis from text \cite{reed2016generative}, super-resolution \cite{ledig2017photo}, style transfer \cite{azadi2018multi}, just to mention a few. Learning useful representation with GANs has shown to heavily rely on hyperparameter-tuning due to various instability issues during training \cite{arjovsky2017towards,grewal2017variance,kurach2018gan}. GANs are remarkably hard to train in spite of their success on variety of task. Robustly and systematically stabilizing the training of GANs has come in many forms such as selective architectural design \cite{radford2015unsupervised}, matching of intermediate features \cite{salimans2016improved}, and unrolling the optimization of discriminator \cite{metz2017unrolled}.\\
\indent
Many recent advances inspired by either theoretical insights or practical considerations have been attempted in form of regularization and
normalization to address some of the issues associated with training of GANs. Imposing Lipschitz constraint on the discriminator has shown to stabilize the adversarial training and avoid an over-optimization scenario where the discriminator still distinguishes and allots different scores to nearly indistinguishable samples \cite{arjovsky2017towards}. By satisfying the Lipschitz constraint, the discriminator's joint/compressed representation of the true and synthesized data distributions is guaranteed to be smooth; thus ensuring a non-zero learning signal for the generator \cite{arjovsky2017towards,grewal2017variance}. Enforcing the discriminator to satisfy the Lipschitz constraints has been approximated and implemented via ancillary means such as gradient penalties \cite{gulrajani2017improved} and weight clipping \cite{arjovsky2017towards}. Using a Gaussian classifier over the real/fake indicator variables has also been shown to have a smoothing effect on the discriminator function \cite{grewal2017variance}.\\
\indent
Injecting label noise \cite{salimans2016improved} and gradient penalty have equally been shown to have a tremendous regularizing effect on GANs. Schemes such as weighted gradient \cite{roth2017stabilizing} and missing modes penalty \cite{che2016mode} have been utilized to alleviate some training and missing modes issues in GAN learning. Techniques such as batch normalization \cite{ioffe2015batch} and layer normalization \cite{ba2016layer} have also been reported in context of GANs \cite{gulrajani2017improved,radford2015unsupervised,denton2015deep}. In batch normalization, pre-activations of nodes in a layer are normalized to mean $\beta$ and standard deviation $\gamma$. Parameters $\beta$ and $\gamma$ are learned for each node in the layer and normalization is done on the batch level and for each node separately \cite{ba2016layer,kurach2018gan}. Layer normalization on the other hand uses the same learned parameters $\beta$ and $\gamma$ to normalize all nodes in a layer and normalizes different samples differently \cite{ba2016layer}. \\
\indent
Weight vectors of discriminator have been $l_2$-normalized with Frobenius norm, which constraints the sum of the squared singular values of the weight matrix to be 1 \cite{salimans2016improved}. However, normalizing using Frobenius norm translates to utilizing a single feature to discriminate the model probability distribution from the target thus, reducing the rank and hence the number of discriminator features \cite{miyato2018spectral}. In addition to weight clipping \cite{arjovsky2017wasserstein,arjovsky2017towards}, weight normalization approaches yield primitive discriminator model that maps the target distribution only with select few features. The most closely related work to ours is orthonormal regularization of weights \cite{brock2016neural} that sets all the singular values of weight matrix in the discriminator to one, which translates to using as many features as possible to distinguish the generator distribution from the target distribution. Our approach, however, imposes much softer orthogonality constraint on the weight vectors by allowing a degree of feature sharing in upper layers of the discriminators. Other related work is spectral normalization of weights that guarantees 1-Lipschitzness for linear layers and ReLu activation units resulting in discriminators of higher rank \cite{miyato2018spectral}. The advantage of spectral normalization is that weight matrices are constrained and Lipschitz. However, bounding the spectral norm of the convolutional kernel to 1 does not bound the spectral norm of the convolutional mapping to unity.
\section{Method}
The training of GAN can be abstracted as a non-cooperative game between two players, namely the generator $G$ and the discriminator $D$. The discriminator tries to distinguish if the generated sample is from the real ($p_{data}$) or fake data distribution ($p_z$), while $G$ tries to trick $D$ into believing that generated sample is from $p_{data}$ by moving the generation manifold towards the data manifold. The discriminator aims to maximize $\E_{\mathbf{x}\sim p_{data}(\mathbf{x})}[log D(\mathbf{x})]$ when the input is sampled from real distribution and given a fake image sample $G(\mathbf{z})$, $\mathbf{z}\sim p_{z}(\mathbf{z})$, it is trained to output probability, $D(G(\mathbf{z}))$, close to zero by maximizing $\E_{\mathbf{z}\sim p_{z}(\mathbf{z})}[log(1-D(G(\mathbf{z})))]$. The generator network, however, is trained to maximize the chances of D producing a high probability for a fake image sample $G(\mathbf{z})$ thus by minimizing $\E_{\mathbf{z}\sim p_{z}}[log(1-D(G(\mathbf{z})))]$.\\
\indent
The adversarial cost is obtained by combining the objectives of both D and G in a min-max game as given in \ref{MyEqGan} below:
\begin{equation}\label{MyEqGan}
 \begin{split}
J_{adv} &= \min_G\max_D\E_{\mathbf{x}\sim p_{data}(\mathbf{x})}[log D(\mathbf{x})] \\
 &+ \E_{\mathbf{z}\sim p_{z}(\mathbf{z})}[log(1-D(G(\mathbf{z})))]
 \end{split}
\end{equation}
Training $D$ can be conceived as training an evaluation metric on sample space \cite{che2016mode} that enables $G$ to use the local gradient $\nabla \log D(G(\mathbf{z}))$ information made available by $D$ to improve itself and move closer to the data manifold.
\subsection{Feature diversification in GAN}
Both D and G are commonly parameterized as DNNs and over the past few years, the general trend has been that DNNs have grown deeper, amounting to huge increase in number of parameters. The number of parameters in DNNs is usually very large offering possibility to learn very flexible high-performing models \cite{liu2014pruning}. Observations from many previous studies \cite{xie2015generalization,rodriguez2016regularizing,dundar2015convolutional,ayinde2018building} suggest that layers of DNNs typically rely on many redundant filters that can be either shifted version of each other or be very similar with little or no variations. For instance, this redundancy is evidently pronounced in filters of AlexNet \cite{krizhevsky2012imagenet} as emphasized in \cite{rodriguez2016regularizing,zeiler2014visualizing,ayinde2016clustering}. To address this redundancy problem, we train layers of the discriminator under specific and well-defined diversity constraints.\\
\indent
Since $G$ and $D$ rely on many redundant filters, we regularize them during training to provide more stable gradient to update both $G$ and $D$. Our regularizer enforces constraints on the learning process by simply encouraging diverse filtering and discourages $D$ from extracting redundant filters. We remark that convolutional filtering has found to greatly benefit from diversity or orthogonality of filters because it can alleviate problems of gradient vanishing or exploding \cite{ayinde2019regularizing,brock2016neural,saxe2013exact,ayinde2017nonredundant}.\\
\indent
Typically, both $D$ and $G$ consist of input, output, and many intermediate processing layers. By letting the number of channels, height, and width of input feature map for  $l^{th}$ layer be denoted as $n_l$, $h_l$, and $w_l$, respectively. A convolutional layer in both $D$ transforms input $\mathbf{x}_l \in \mathbb{R}^{p}$ into output $\mathbf{x}_{l+1} \in \mathbb{R}^{q}$, where $\mathbf{x}_{l+1}$ is the input to layer $l+1$;  $p$ and $q$ are given as $n_l\times h_l\times w_l$ and $n_{l+1}\times h_{l+1}\times w_{l+1}$, respectively. $\mathbf{x}_l$ is convolved with $n_{l+1}$ $3D$ filters $\chi \in \mathbb{R}^{n_l \times k\times k}$, resulting in $n_{l+1}$ output feature maps. Unrolling and combining all layer $l^{th}$ filters into a single matrix results in kernel matrix $\overset{(l)}{\Theta^D} \in \mathbb{R}^{m\times n_{l+1}}$ where $m= k^2n_l$. Then, $\overset{(l)}{\theta^D}_i$, i=1,...$n_l$, denotes filters in layer $l$, each $\overset{(l)}{\theta^D}_i \in \mathbb{R}^{m}$ corresponds to the $i$-th column of the kernel matrix $\overset{(l)}{\Theta^D} = [\overset{(l)}{\theta^D}_1, \;\;...\overset{(l)}{\theta^D}_{n_l}] \in \mathbb{R}^{m\times n_{l+1}}$; the bias term of each layer is omitted for simplicity.
Given that $\overset{(l)}{\Theta^D} \in \mathbb{R}^{m\times n_l}$ contain $n_l$ normalized filter vectors as columns, each with $m$ elements corresponding to connections from layer $l-1$ to $i^{th}$ neuron of layer $l$, then, the diversity loss $J_D$ for all layers of $D$ is given as:
\begin{equation} \label{MyEq1}
  \begin{split}
 J_D(\theta^D)=\sum_{l=1}^{L}\left(\frac{1}{2}\sum_{i=1}^{n_l}\sum_{j=1}^{n_l}\left(\overset{(l)}{\Omega_{ij}^D}\right)^{2}\overset{(l)}{\mathbf{M}_{ij}^D}\right)
  \end{split}
\end{equation}
where $\overset{(l)}{\Omega^D} \in \mathbb{R}^{n_l\times n_l}$ denotes $(\overset{(l)}{\Theta^D})^T\overset{(l)}{\Theta^D}$ which contains the inner products of each pair of columns $i$ and $j$ of $\overset{(l)}{\Theta^D}$ in each position $i$,$j$ of $\overset{(l)}{\Omega^D}$ in layer $l$; $\overset{(l)}{\textbf{M}^D} \in \mathbb{R}^{n_l\times n_l}$ is a binary mask for layer $l$ defined in \eqref{MyEq2}; $L$ is the number of layers to be regularized.
\begin{equation}\label{MyEq2}
\overset{(l)}{\mathbf{M}_{ij}^D} = \Bigg\{
\begin{array}{l l}
1 & \quad \tau \leq |\overset{(l)}{\Omega_{ij}^D}| \leq 1 \\
0 & \quad i = j \\
0 & \quad otherwise
\end{array}
\end{equation}
Similarly, the diversity loss $J_G$ for generator $G$ is given as:
\begin{equation} \label{MyEq1}
  \begin{split}
 J_G(\theta^G)=\sum_{l=1}^{L}\left(\frac{1}{2}\sum_{i=1}^{n_l}\sum_{j=1}^{n_l}\left(\overset{(l)}{\Omega_{ij}^G}\right)^{2}\overset{(l)}{\mathbf{M}_{ij}^G}\right)
  \end{split}
\end{equation}
and
\begin{equation}\label{MyEq2}
\overset{(l)}{\mathbf{M}_{ij}^G} = \Bigg\{
\begin{array}{l l}
1 & \quad \tau \leq |\overset{(l)}{\Omega_{ij}^G}| \leq 1 \\
0 & \quad i = j \\
0 & \quad otherwise
\end{array}
\end{equation}
It is important to also note the importance and relevance of $\tau$ in \eqref{MyEq2}. Setting $\tau=0$ results in layer-wise disjoint filters. This forces weight vectors to be orthogonal by pushing them towards the nearest orthogonal manifold. However, from practical standpoint, disjoint filters are not desirable because some features are sometimes required to be shared with layers. For instance a model trained on CIFAR-10 dataset \cite{krizhevsky2009learning} that have "automobiles" and "trucks" as two of its ten categories, if a particular lower-level feature captures "wheel" and two higher-layer features describe automobile and truck, then it is highly probable that the two upper layer features might share the feature that describe the wheel. The choice of $\tau$ determines the level of sharing allowed, that is, the degree of feature sharing across features of a particular layer. In other words, $\tau$ serves as a trade-off parameter that ensures some degree of feature sharing across multiple high-level features and at the same time ensuring features are sufficiently dissimilar.\\
\indent
\begin{figure*}[htb!]
	\begin{minipage}[b]{0.5\linewidth}%
		\centering
		\centerline{\includegraphics[scale=0.45]{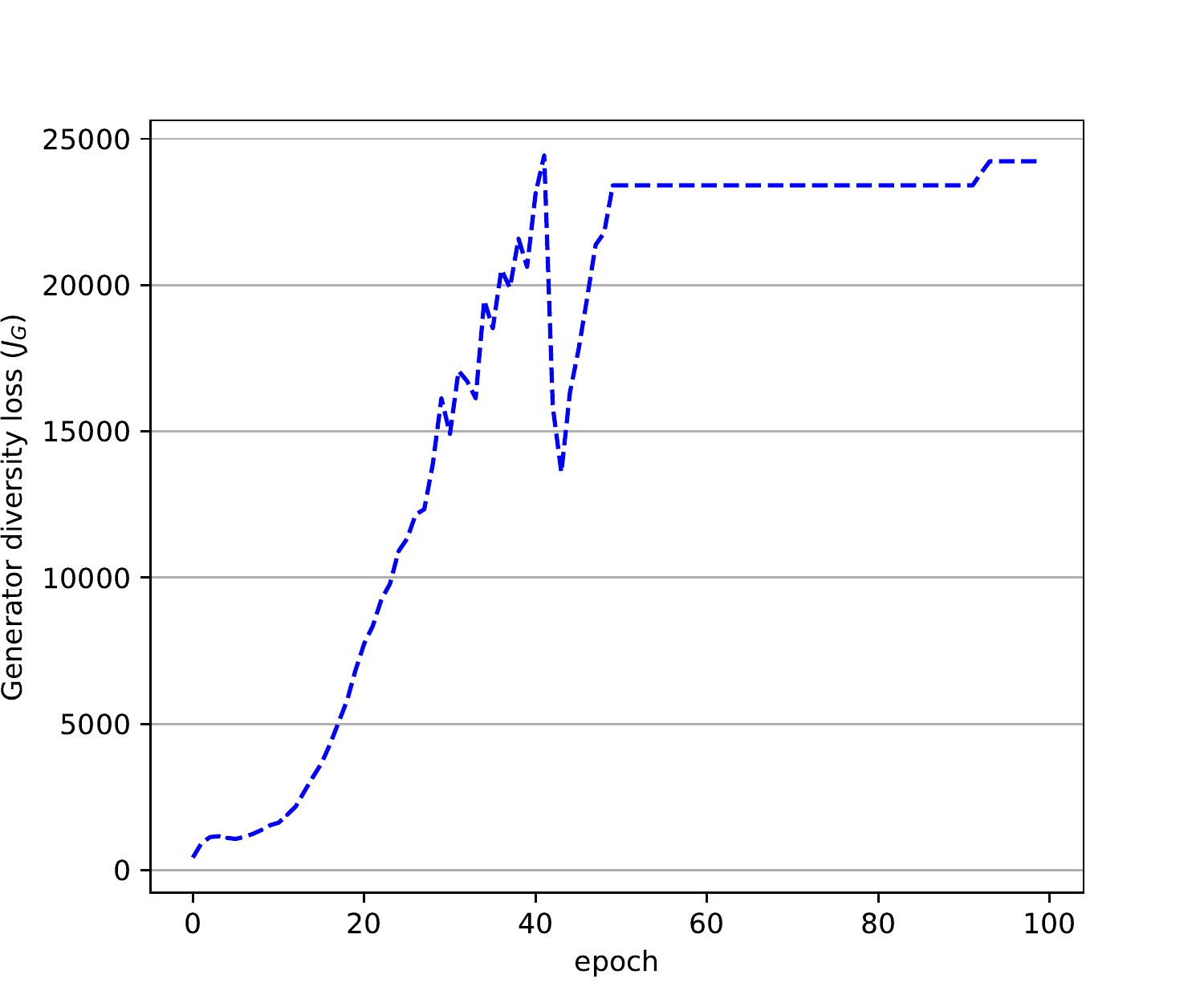}}
		{{\footnotesize (a)}}
	\end{minipage}
\begin{minipage}[b]{0.5\linewidth}
		\centering
		\centerline{\includegraphics[scale=0.45]{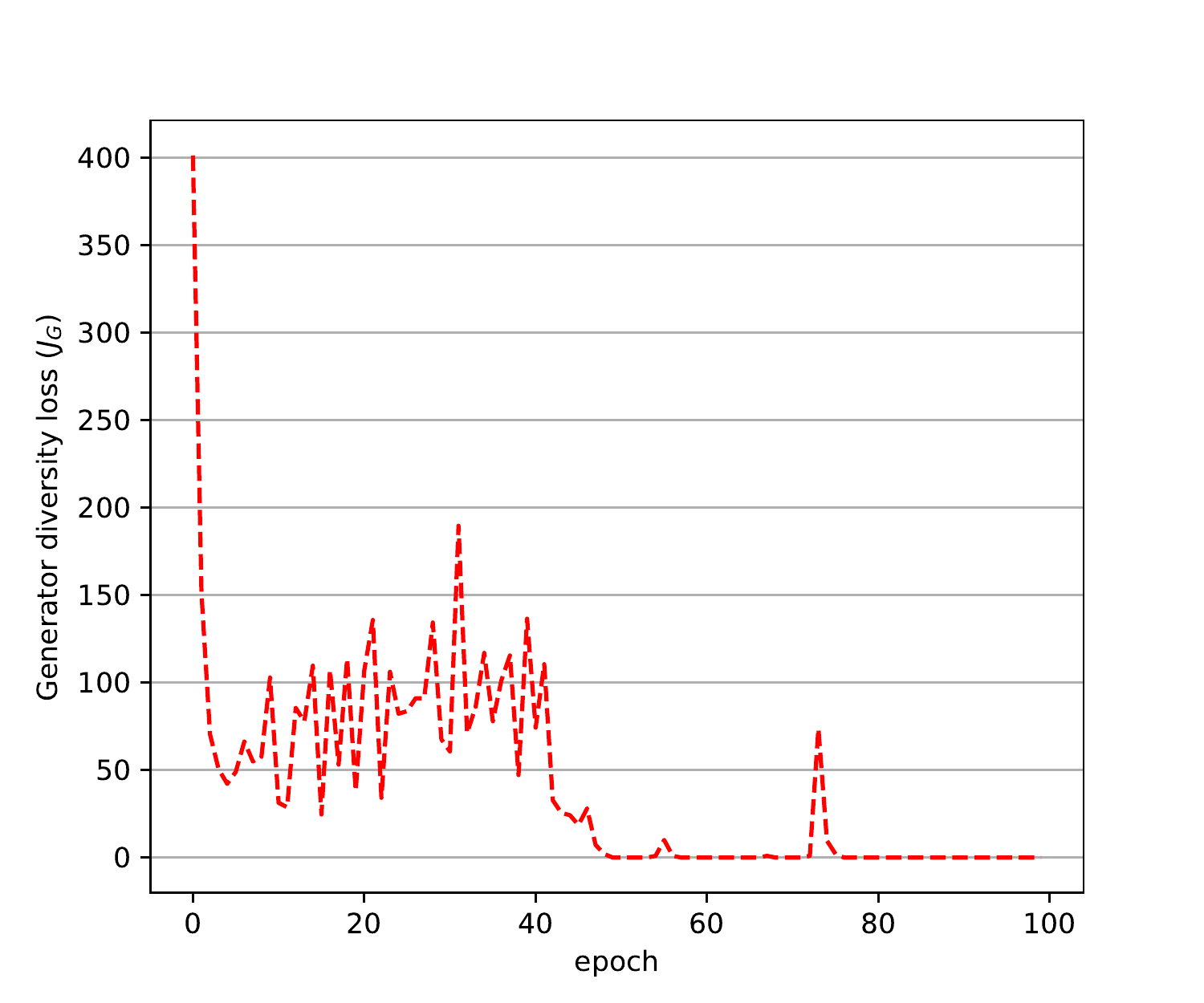}} %
		{{\footnotesize(b)}}
	\end{minipage}
\begin{minipage}[b]{0.5\linewidth}
		\centering
		\centerline{\includegraphics[scale=0.45]{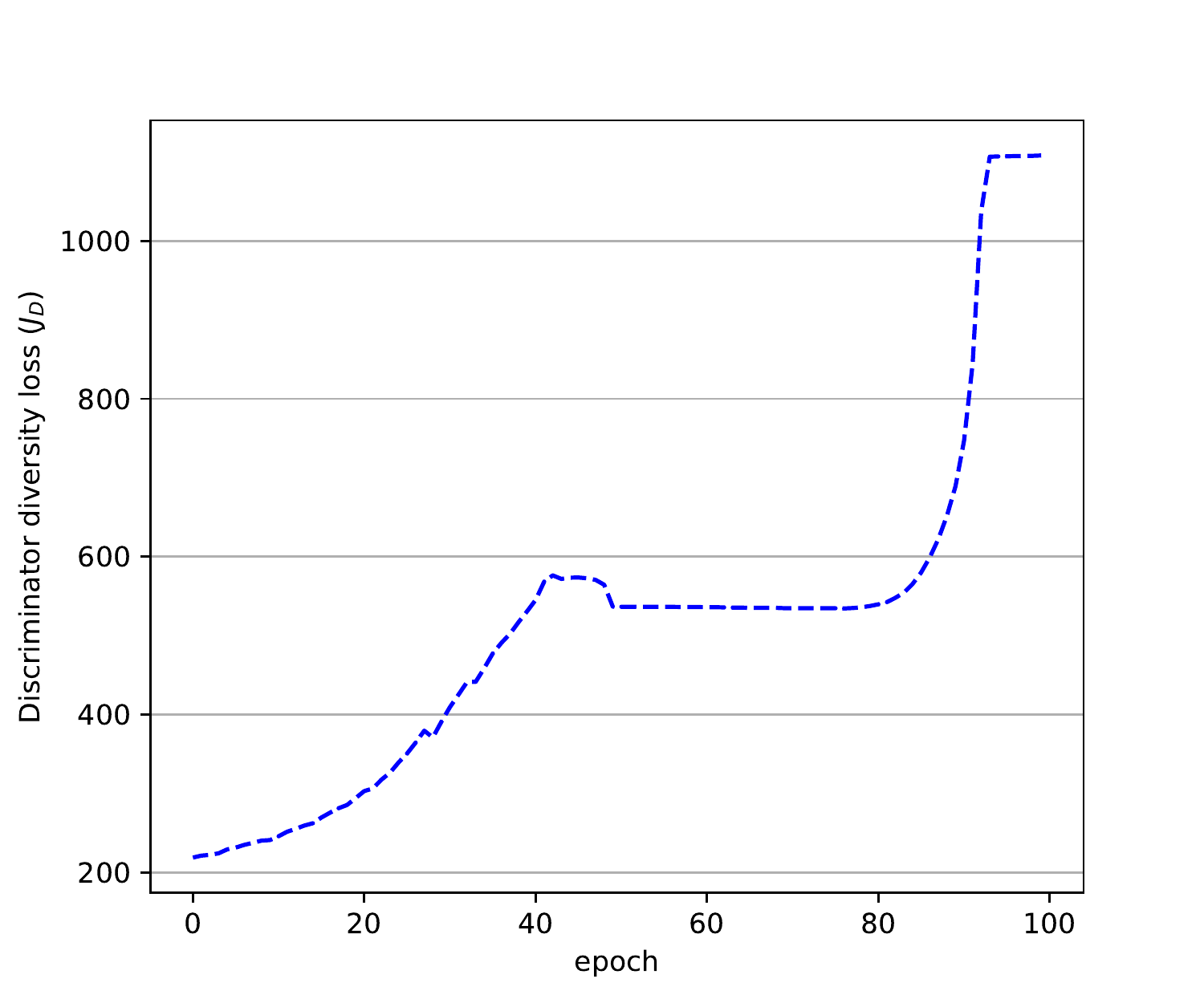}} %
		{{\footnotesize(c)}}
	\end{minipage}
\begin{minipage}[b]{0.5\linewidth}
		\centering
		\centerline{\includegraphics[scale=0.45]{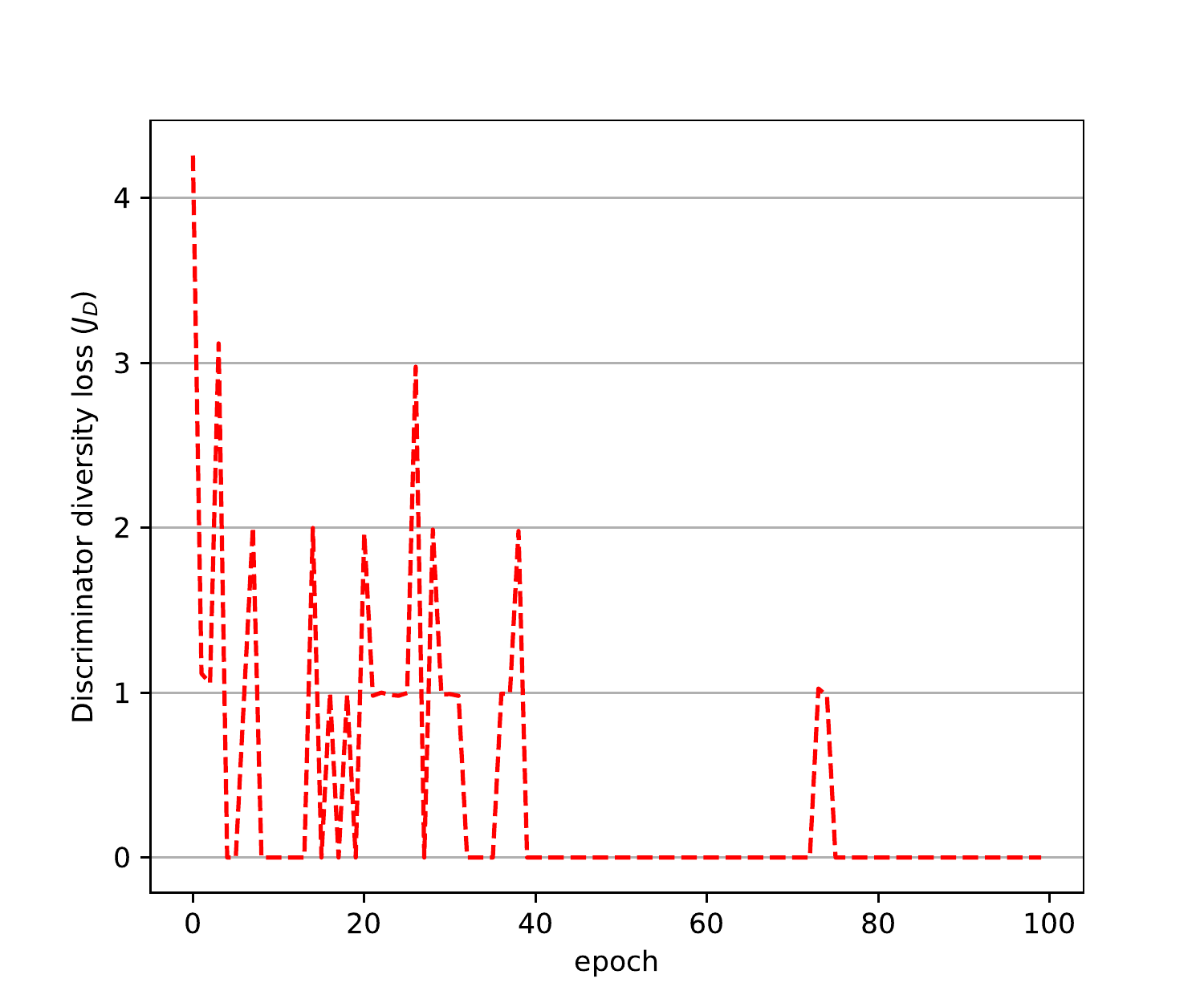}} %
		{{\footnotesize(d)}}
	\end{minipage}
	\caption{Diversity loss of (a) generator $J_G$ with no regularization (b) generator $J_G$ with diReAL (c) discriminator $J_D$ with no regularization, and (d) discriminator $J_D$ with DiReAL trained on MNIST dataset.}\label{div_loss_mnist}
\end{figure*}
In order to enforce feature diversity in both $G$ and $D$ while training GANs, the diversity regularization terms in \eqref{MyEq1} is added to the conventional adversarial cost $J_{adv}$ in \eqref{MyEqGan} as given in \eqref{MyEqGan2}.\\
\begin{equation}\label{MyEqGan2}
 \begin{split}
J_{net} &= J_{adv} + J_{div}
 \end{split}
\end{equation}
where  $J_{div} = \lambda_G J_G(\theta^G) - \lambda_D J_D(\theta^D)$, $\lambda_G$ and $\lambda_D$ is the diversity penalty factors for generator and discriminator, respectively. The derivative of diversity loss $J_D$ with respect to weights of $D$ is given as
\begin{equation}\label{MyEq9bb}
\begin{split}
\nabla_{\Theta_{i,j}^{(l)}}J_D(\theta^D) = \sum_{k=1}^{n} \overset{(l)}{\Theta_{i,k}^D} \overset{(l)}{\Omega_{k,j}^D} \overset{(l)}{\mathbf{M}_{k,j}^D}
\end{split}
\end{equation}
and the derivative of diversity loss $J_G$ with respect to weights of $G$ is
\begin{equation}\label{MyEq9bb}
\begin{split}
\nabla_{\Theta_{i,j}^{(l)}}J_G(\theta^G) = \sum_{k=1}^{n} \overset{(l)}{\Theta_{i,k}^G} \overset{(l)}{\Omega_{k,j}^G} \overset{(l)}{\mathbf{M}_{k,j}^G}
\end{split}
\end{equation}
The idea behind diversifying features is that in addition to adversarial gradient information provided by $D$, we provide additional diversity loss with more stable gradient to refine both $G$ and $D$. The diversity loss encourages weights of both generator and discriminator to be diverse by pushing them towards the nearest orthogonal manifold. Our proposed regularization provides more efficient gradient flow, a more stable optimization, richness of layer-wise features of resulting model, and improved sample quality compared to benchmarks and baseline. The diversity regularization ensures the column space of $\overset{(l)}{\Theta^D}$ and $\overset{(l)}{\Theta^G}$ for $l^{th}$ layer does not concentrate in few direction during training thus preventing them to be sensitive in few and limited directions. The proposed diversity regularized adversarial learning alleviates some of the main failure mode of GAN by ensuring features are diverse.
\section{Experiments}
\begin{figure}[htb!]
	\begin{minipage}[b]{0.495\linewidth}
		\centering
		\centerline{\includegraphics[scale=0.45]{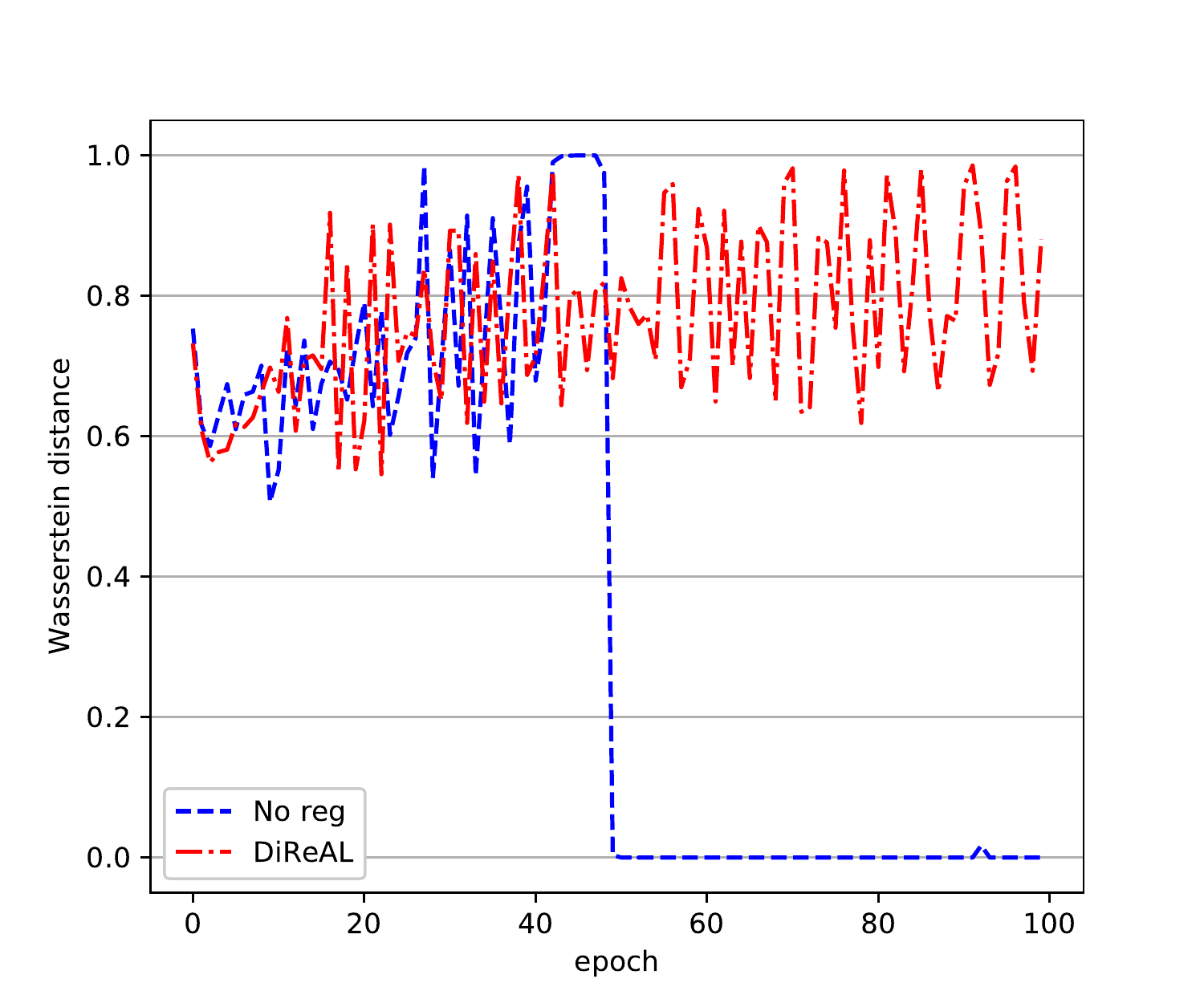}}
		{{\footnotesize (a)}}
	\end{minipage}
\begin{minipage}[b]{0.495\linewidth}
		\centering
		\centerline{\includegraphics[scale=0.45]{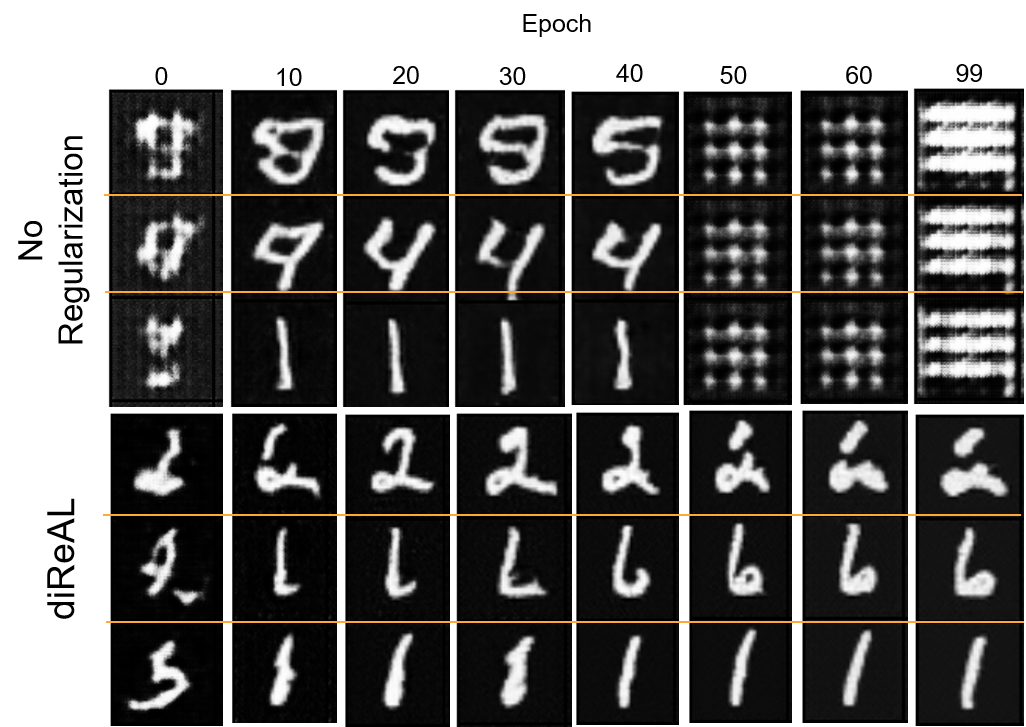}} %
		{{\footnotesize(b)}}
	\end{minipage}
	\caption{(a) Divergence, as measured by Wasserstein distance, between the discriminator output for synthesized and real MNIST samples (b) Synthesized hand-written digits with and without diversity regularization.}\label{mnist_wa_dist}
\end{figure}

All experiments were performed on Intel(r) Core(TM) i7-6700 CPU @ 3.40Ghz and a 64GB of RAM running a 64-bit Ubuntu 16.04 edition. The software implementation has been in PyTorch library \footnote{https://pytorch.org/} on two Titan X 12GB GPUs. Implementation of DiReAL will be available at \url{https://github.com/babajide07/DiReAL-PyTorch-Implementation}. Diversity regularized adversarial learning (DiReAL) was evaluated on MNIST dataset of handwritten digits\cite{lecun1998mnist}, CIFAR-10 \cite{krizhevsky2009learning}, STL-10 \cite{coates2011analysis}, and Celeb-A \cite{liu2015deep} databases.  In the first set of experiments, an ubiquitous deep convolutional GAN (DCGAN) in \cite{radford2015unsupervised} was trained using MNIST digits. The standard MNIST dataset has 60000 training and 10000 testing examples. Each example is a grayscale image of an handwritten digit scaled and centered in a 28 $\times$ 28 pixel box. Both the discriminator and generator networks contain 5 layers of convolutional block. Adam optimizer \cite{kingma2014adam} with batch size of 64 was used to train the model for 100 epochs and $\tau$ and learning rate in DiReAL were set to 0.5 and $0.0001$, respectively. In similar vein, $\lambda_D$ and $\lambda_G$ were to 1.0 and 0.01, respectively. Adam optimizer ($\beta_1=0.0$, $\beta_2=0.9$) \cite{kingma2014adam} with batch size of 64 was used to train the model for 100 epochs\\
\indent
Fig.~\ref{div_loss_mnist} shows the diversity loss of both generator and discriminator for DiReAL and unregularized counterpart. It can be observed that DiReAL was able to minimize the pairwise feature correlations compared to the highly correlated features extracted by the unregularized counterpart. Specifically, DiReAL was able to steadily minimize the diversity loss as training progresses compared to the unregularized DCGAN, where extraction of similar features grows with epoch of training, thus increasing the diversity loss. The divergence between discriminator output for real handwritten digits and generated samples over 30 batches for regularized and the unregularized networks is shown in Fig.~\ref{mnist_wa_dist}a. The divergence was measured using the Wasserstein distance measure \cite{vallender1974calculation} and it can be observed that the regularizing effect of DiReAL stabilizes the adversarial training and prevents mode collapse. For unregularized network, however, the mode started to collapse around 45th epoch. Closer look into the diversity of the generator in Fig.~\ref{div_loss_mnist}a, it is evident that just around the epoch of collapse the generator starts extracting more and more redundant filters. We suspect that DiReAL was able to stabilize the training by pushing features to lie close to the orthogonal manifold, thus preventing learned features from collapsing to an undesirable manifold. Fig.~\ref{mnist_wa_dist}b shows the handwritten digit samples synthesized with and without DiReAL and it can be observed that diversification of features is beneficial for stabilizing adversarial learning and ultimately improving the samples' quality. Another observation is that DiReAL also prevents learned weights from collapsing to an undesirable manifold thus highlighting some of the benefits of pushing weights near the orthogonal manifold.\\
\indent
\begin{figure}[htb!]
	\begin{minipage}[b]{0.495\linewidth}
		\centering
		\centerline{\includegraphics[scale=0.45]{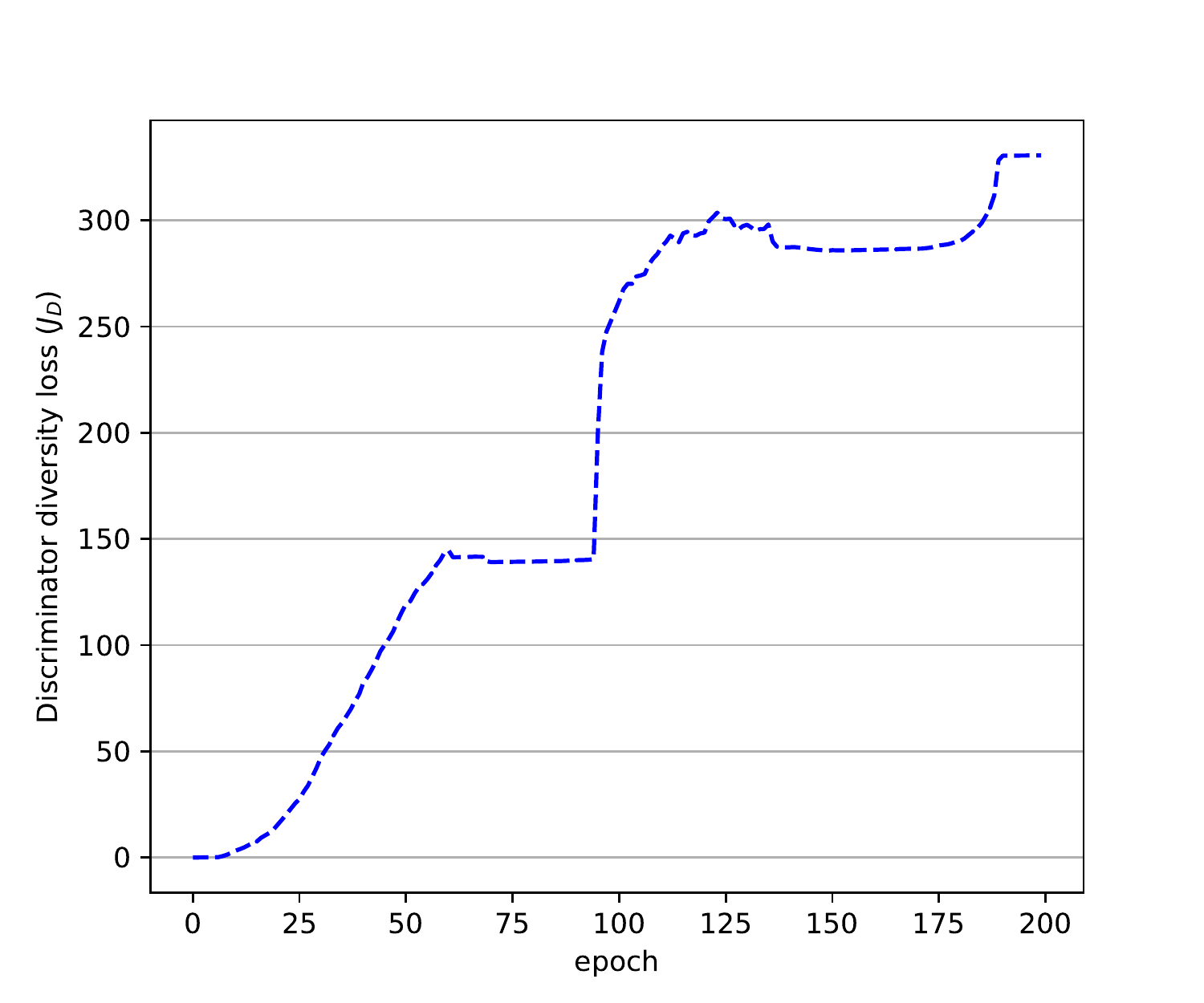}}
		{{\footnotesize (a)}}
	\end{minipage}
\begin{minipage}[b]{0.495\linewidth}
		\centering
		\centerline{\includegraphics[scale=0.45]{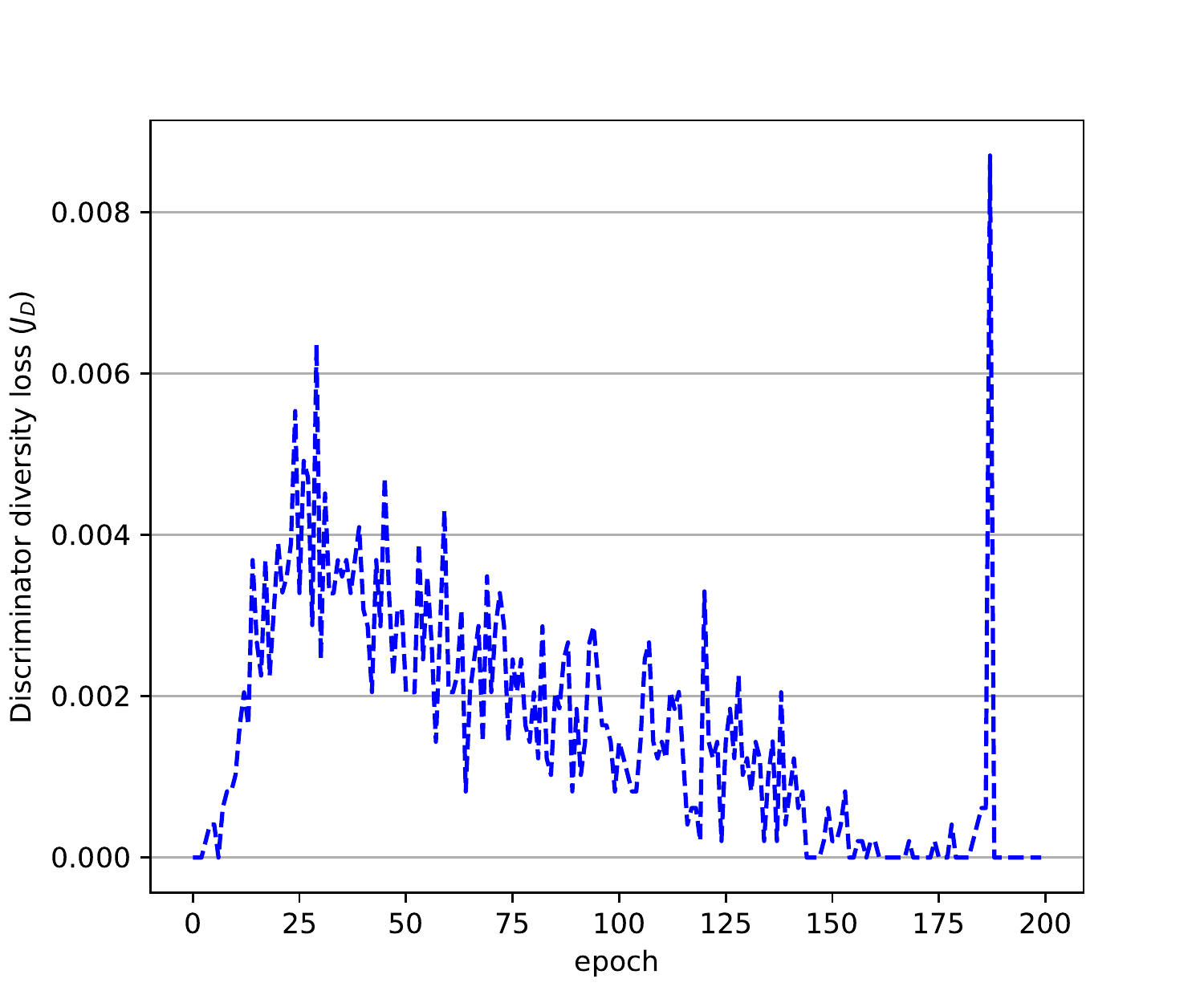}} %
		{{\footnotesize(b)}}
	\end{minipage}
	\caption{Diversity loss of (a) discriminator $J_D$ with no regularization, and (b) discriminator $J_G$ with diReAL trained on CIFAR-10 dataset.}\label{cifar_divloss}
\end{figure}
\begin{figure*}[htb!]
    \begin{minipage}[b]{1.0\linewidth}
		\centering
       \captionsetup{justification=centering}
		\centerline{\includegraphics[scale=0.5]{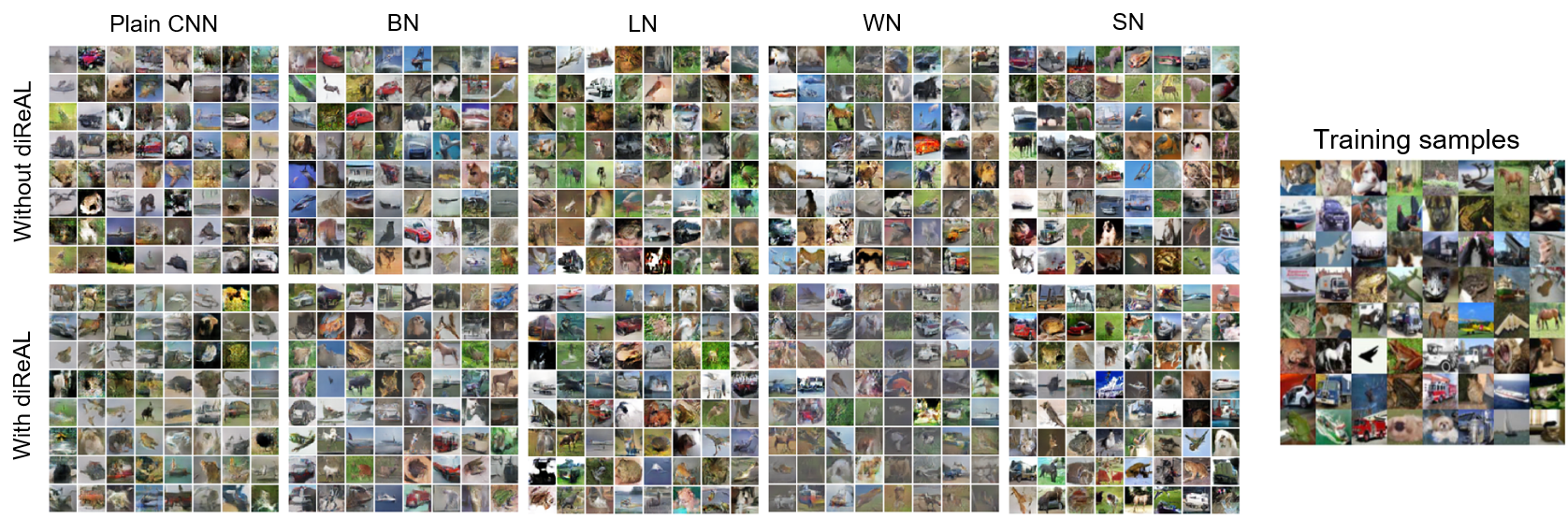}}
	\end{minipage}
	\caption{Generated images with and without DiReAL trained on CIFAR-10 dataset.}\label{cifar10_select}		
\end{figure*}
\begin{table}\centering
\scalebox{1.2}{
\begin{tabular}{l|c}
\hline
Method & Inception Score \\
\hline
Real data & 9.04 \\
\hline
\textbf{-Standard CNN-} &  \\
Unregularized \cite{radford2015unsupervised}& 4.00 $\pm$ 0.15 \\
DiReAL (ours) & 4.17 $\pm$ 0.03 \\
Batch Normalization \cite{ioffe2015batch} & 5.48 $\pm$ 0.19 \\
Layer Normalization \cite{ba2016layer} & 5.05 $\pm$ 0.12 \\
Weight Normalization \cite{salimans2016weight} & 4.66 $\pm$ 0.14 \\
Spectral Normalization \cite{miyato2018spectral} & \textbf{6.50} $\pm$ 0.30 \\
\hline
Weight Normalization + DiReAL & 4.68 $\pm$ 0.06\\
Batch Normalization + DiReAL & 5.48 $\pm$ 0.15 \\
Layer Normalization + DiReAL & 5.64 $\pm$ 0.15 \\
Spectral Normalization + DiReAL & \textbf{6.87} $\pm$ 0.12 \\
\hline
\end{tabular}}
\caption{Inception Scores with unsupervised image generation on CIFAR-10}\label{inception_cifar10}
\end{table}
\begin{figure*}[htb!]
    \begin{minipage}[b]{1.0\linewidth}
		\centering
       \captionsetup{justification=centering}
		\centerline{\includegraphics[scale=0.5]{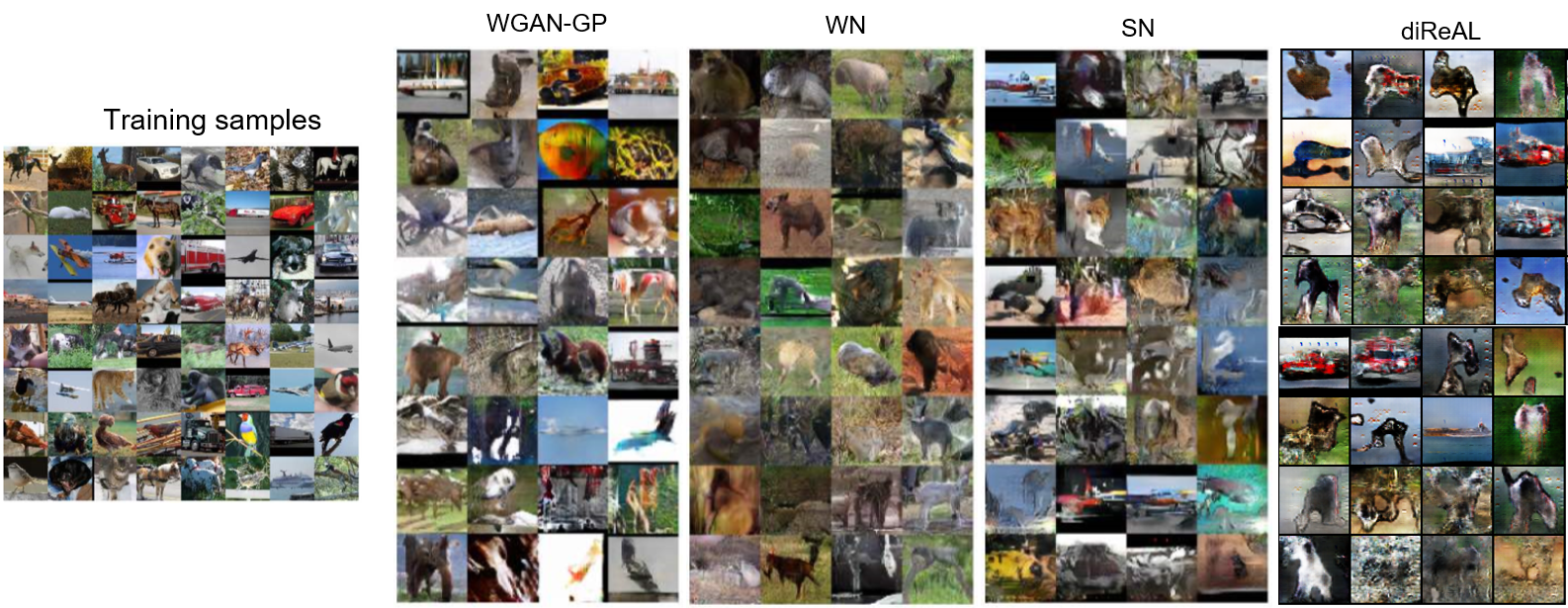}}
	\end{minipage}
	\caption{Qualitative comparison of generated images with four regularization techniques for models trained on STL-10 dataset.}\label{stl10_select}		 \end{figure*}
\begin{figure}[htb!]
    \begin{minipage}[b]{1.0\linewidth}
		\centering
       \captionsetup{justification=centering}
		\centerline{\includegraphics[scale=0.9]{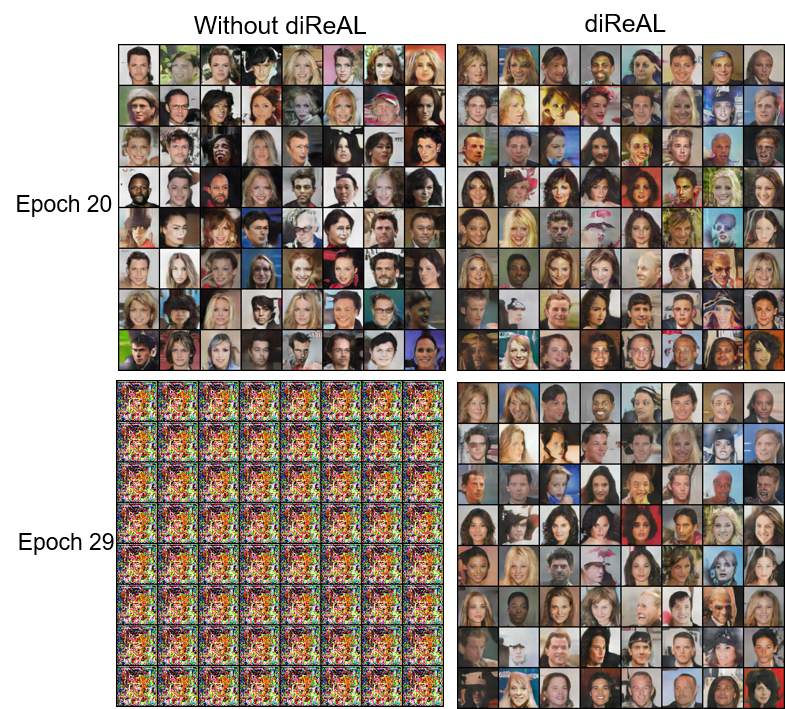}}
	\end{minipage}
	\caption{Generated images with and without diversity Regularization trained on CELEB-A dataset.}\label{celeba_select}		
\end{figure}
In the second large-scale experiments, CIFAR-10 dataset was used to train GAN using DiReAL and the results compared to the unregularized training. The dataset is split into $50000$ and $10000$ training and testing sets, respectively. Similar to experiments with MNIST, Fig.~\ref{cifar_divloss}b shows the diversity loss of the discriminator with and without DiReAL trained on CIFAR-10 database. It can be observed that DiReAL was able to minimize the diversity loss and encourages diverse features that benefit the adversarial training. On the other hand, Fig.~\ref{cifar_divloss}b shows that the diversity loss of the unregularized is higher and unconstrained compared to that of DiReAL.  The images synthesized with DiReAL was compared and contrasted with state-of-the-art methods such as batch normalization \cite{ioffe2015batch}, layer normalization \cite{ba2016layer}, weight normalization \cite{salimans2016weight}, and spectral normalization \cite{miyato2018spectral}. It is remarked that DiReAL can be used in tandem with the other regularization techniques and could also be deployed as stand-alone regularization tool for stabilizing adversarial learning. In this light, DiReAL was also combined with these techniques. It must be noted that spectral normalization uses a variant of DCGAN architecture with an eight-layer discriminator network. See \cite{miyato2018spectral} for more implementation details.\\
\indent
It can be observed in Fig.~\ref{cifar10_select} that diversity regularization was able to synthesize more diverse and complex images compared to unregularized counterpart. Other benchmark regularizers were able to generate better image samples compared to using only DiReAL. However, when DiReAL was combined with other regularizers the quality of the generated samples was significantly improved. For quantitative evaluation of generated examples, inception score metric \cite{salimans2016weight} was used. Inception score has been found to highly correlate with with subjective human judgment of image quality \cite{salimans2016weight,miyato2018spectral}. Similar to \cite{salimans2016weight,miyato2018spectral}, inception score was computed for 5000 synthesized images using generators trained with each regularization technique. Every run of the experiment is repeated five times and averaged to combat the effect of random initialization. The average and the standard deviation of the inception scores are reported.
\\
\indent
The proposed regularization is also compared and contrasted in terms inception score with many benchmark methods as summarized in Table~\ref{inception_cifar10}. It can be again observed that DiReAL was able to improve the image generation quality compared to unregularized counterpart and when combined with spectral normalization, we observed a 6\% improvement in the inception score. By combining DiReAL with layer normalization, an improvement of 11.68\% on inception was observed. However, no significant improvement was observed when DiReAL was combined with batch normalization and weight normalization. It must be remarked that the calculation of Inception Scores is library dependent and that is why the scores reported in Table~\ref{inception_cifar10} is different for those reported by Miyato et al. \cite{miyato2018spectral}. While our implementation was in PyTorch, \cite{miyato2018spectral} was in Chainer \footnote{https://chainer.org/}.
\\
\indent
In the next set of large-scale experiments, STL-10 dataset was used to train generator under diversity regularization and compared with other state-of-the-art regularization techniques. As can be observed in Fig.~\ref{stl10_select}, images synthesized by generator trained with DiReAL was able to generate images with competitive quality in comparison with other regularization methods considered. Performance of DiReAL was also observed to be competitive to regularization methods such as WGAN-GP and spectral normalization. In Fig.~\ref{celeba_select} we show the images produced by the generators trained with DiReAL using Celeb-A dataset. It can be again be observed that DiReAL was able to stabilize the training and avoid mode collapse in comparison to the unregularized counterpart.
\section{Conclusion}
This paper proposes a good method of stabilizing the training of GANs using diversity regularization to penalize both negatively and positively correlated features according to features differentiation and based on features relative cosine distances. It has been shown that diversity regularization can help alleviate a common failure mode where the generator collapses to a single parameter configuration and outputs identical points. This has been achieved by providing additional stable diversity gradient information in addition to adversarial gradient information to update both the generator and discriminator's features. The performance of the proposed regularization in terms of extracting diverse features and improving adversarial learning was compared on the basis of image synthesis with recent regularization techniques namely batch normalization, layer normalization, weight normalization, weight clipping, WGAN-GP, and spectral normalization. It has also been shown on select examples that extraction of diverse features improves the quality of image generation, especially when used in combination with spectral normalization. This concept is illustrated using MNIST handwritten digits, CIFAR-10, STL-10, and Celeb-A Dataset.
\bibliographystyle{IEEEtran}
\bibliography{egbib}
\end{document}